\documentclass[11pt]{article}

\usepackage[preprint]{acl}

\usepackage{times}
\usepackage{latexsym}

\usepackage[T1]{fontenc}
\usepackage[utf8]{inputenc}

\usepackage{microtype}

\usepackage{inconsolata}

\usepackage{graphicx}
\usepackage{amsmath, amssymb, amsthm}
\usepackage{graphicx}
\usepackage{hyperref}
\usepackage{algorithm}
\usepackage{algorithmic}
\usepackage{fullpage}
\usepackage{booktabs}
\usepackage{multirow}
\usepackage[table]{xcolor}
\definecolor{lightgreen}{RGB}{225,245,225}

\usepackage{comment}

\usepackage{tabularx}
\usepackage{array}

\definecolor{table_green}{RGB}{34,139,34}

\title{Representation-Aware Advantage Estimation: \\Your Reward Model Provides More Than A Scalar Output}

\author{Guozheng Li\thanks{Equal contribution.}\\
  Southeast University \\
  \texttt{gzli@seu.edu.cn} \\\And
  Xiyan Fu\footnotemark[1] \\
  Nanyang Technological University \\
  \texttt{xiyan.fu@ntu.edu.sg} \\\And
  Yiwen Guo\thanks{Corresponding author.} \\
  Independent Researcher \\
  \texttt{guoyiwen89@gmail.com} \\
  }

\begin{document}
\maketitle
\begin{abstract}
Current reinforcement learning from human feedback (RLHF) methods primarily rely on scalar rewards from a trained reward model (RM). While effective, scalar rewards are often noisy and fail to capture fine-grained preference differences, whereas RM hidden states encode richer semantic and preference information. We introduce the representation-aware advantage estimation, which leverages RM hidden states and models them as auxiliary signals for better advantage estimation. Specifically, we propose the Graph-based Advantage Estimation (GraphAE), treat each sampled group as a graph, where nodes correspond to responses and edges capture their similarity in the RM hidden space. Then advantages are computed via graph propagation, enabling each sample to incorporate contextual information from its neighbors. GraphAE is lightweight and can be seamlessly integrated into existing group-based RL algorithms. We apply GraphAE to GRPO, GSPO and RLOO, and conduct extensive experiments on different models and benchmarks. Empirical results show consistent improvements across three benchmarks, with gains of up to + 6.3 on Arena-Hard-v0.1, + 8.27 on AlpacaEval 2.0, and + 0.22 on MT-Bench. These results demonstrate that leveraging RM representations leads to more sample efficient and robust RLHF~\footnote{\href{https://github.com/1iguozheng/graphae}{https://github.com/1iguozheng/graphae}}.
\end{abstract}

\section{Introduction}
Reinforcement Learning from Human Feedback (RLHF)~\citep{christiano2017deep} has emerged as a cornerstone for aligning Large Language Models (LLMs) with human preferences~\citep{bai2022constitutional,ganguli2022red}. The standard RLHF pipelines rely on a scalar reward produced by a reward model (RM) to guide policy updates via policy gradient methods~\cite{sutton1999policy}. Many policy optimization methods have been developed to improve the stability and efficiency of RLHF, including PPO~\cite{schulman2017proximal}, group-based variants such as GRPO~\cite{shao2024deepseekmath}, as well as direct preference optimization methods such as DPO~\cite{rafailov2023direct}. These algorithms have been widely applied to tasks such as instruction following~\cite{ouyang2022training}, dialogue~\cite{zhang2020dialogpt}, and reasoning~\cite{wei2022chain}.

\begin{figure}[t]
	\centering
	\includegraphics[width=\linewidth]{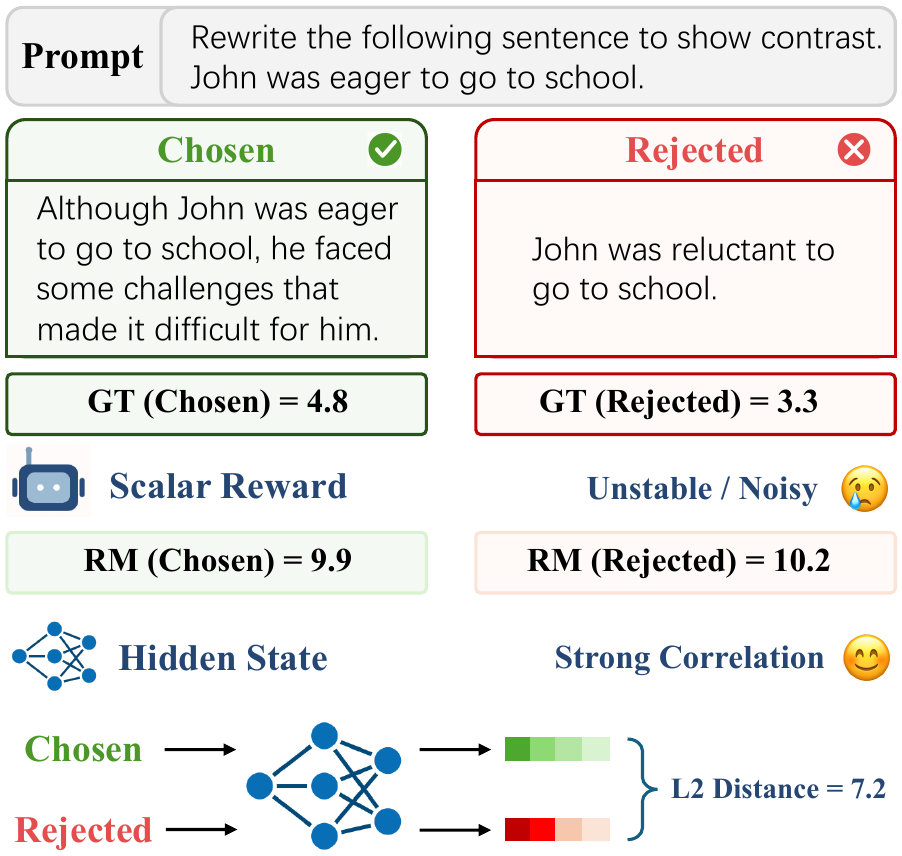}
	\caption{The same RM may assign similar or even reversed scalar scores to opposite responses, while its hidden representations clearly separate them. GT denotes ground truth and RM denotes reward model.}
	\label{motivation}
    \vspace{-4mm}
\end{figure}

Although these methods have achieved strong empirical performance, they heavily rely on the scalar output of the RM. However, we find different responses may receive identical or very similar scores despite exhibiting clear semantic differences, as shown in Figure~\ref{motivation}. That is, a scalar reward is often insufficient to capture fine-grained differences between responses. As a result, the advantage estimates computed in many online RL algorithms based on such reward signals can be inaccurate or noisy~\cite{zhang2025policy}.

To address this limitation, we aim to identify a more expressive signal that can better capture fine-grained differences among responses. Recent studies on interpretable RMs suggest that intermediate representations encode rich, fine-grained information related to preference and reward signals that is not fully reflected in the final scalar output \citep{wang2024interpretable,zhang2026interpretable}. We therefore utilize the RM’s hidden state representations prior to the final scalar projection for advantage estimation. Building on these representations, we move beyond scalar advantage estimation and introduce a Representation-Aware Advantage Estimation.

%Scalar rewards are highly compressed representations, which makes it difficult to preserve such fine-grained distinctions. Instead, we turn to the RM’s hidden-state representations, which are computed prior to the final scalar projection and therefore retain richer information about the underlying responses. Based on this, we leverage hidden-state representations to define a similarity structure over responses, which is then used to propagate and refine the original advantage signal.

% In specific, we propose Graph Advantage Estimation (GraphAE), a structured advantage estimator built upon RM representations. Our key idea is to leverage reward features to construct a graph that captures local correlations among sampled completions, and to compute advantages by propagating rewards over this graph. Given a set of sampled responses, we construct a similarity graph based on their RM representations, where edge weights reflect pairwise similarity. We then estimate advantages by solving a graph-regularized objective that balances fidelity to the original scalar rewards with smoothness over the graph~\citep{smola2003kernels,zhu2003semi}. This yields a closed-form solution corresponding to a graph Laplacian, which can be efficiently solved in practice. The resulting advantages integrate both individual reward signals and local structural information, leading to a more informative and robust estimator.

In specific, we propose the Graph-based Advantage Estimation (GraphAE), a structured advantage estimator built upon RM representations. Given a set of sampled responses, we construct a similarity graph based on their RM representations, where edge weights reflect pairwise similarity. We then estimate advantages by solving a graph-regularized objective that balances fidelity to the original scalar rewards with smoothness over the graph~\citep{smola2003kernels,zhu2003semi}. This yields a closed-form solution that can be efficiently solved in practice. The resulting rewards integrate both individual reward signals and local structural information, leading to a more informative and robust estimator. It is lightweight and can be seamlessly integrated into existing group-based RL algorithms by replacing the original advantage estimator, such as GRPO~\citep{shao2024deepseekmath}, GSPO~\citep{zheng2025group}, and RLOO~\citep{ahmadian2024back}.

We evaluate GraphAE on widely used benchmarks including Arena-Hard-v0.1~\citep{li2024crowdsourced}, AlpacaEval 2.0~\citep{dubois2024length}, and MT-Bench~\citep{zheng2023judging}. GraphAE consistently improves performance over strong baselines on different RL algorithms and backbones, with gains of up to +6.3 on Arena-Hard, +8.27 on AlpacaEval, and +0.22 on MT-Bench, demonstrating improved alignment quality and robustness. Moreover, GraphAE reduces within-group reward variance and accelerates convergence. GraphAE is robust across hyperparameter choices and adds minimal computation during advantage estimation, showing that leveraging RM representations leads to more stable and efficient RLHF.

\section{Related Work}

\subsection{Reinforcement Learning for LLMs}
Reinforcement learning has emerged as a prevalent method for fine-tuning LLMs, with PPO~\cite{schulman2017proximal} and its variations~\cite{ramamurthy2023reinforcement,wu2023pairwise} being the most prevalent methods. These techniques largely adhere to the actor-critic paradigm~\cite{sutton1998reinforcement}, which significantly increases training complexity and computational resource consumption. Many recent studies have sought to mitigate this limitation. Methods such as DPO~\cite{rafailov2023direct} and its variants~\cite{meng2024simpo,ethayarajh2024kto,hong2024orpo} bypass reward modeling and the actor-critic learning framework by directly learning from preferences. However, existing research indicates that due to their offline nature, these approaches exhibit a performance gap compared to online RL~\cite{xu2024dpo}. In contrast to actor-critic and offline methods, recent work explores \emph{group-based policy optimization}~\cite{shao2024deepseekmath,yu2025dapo,liu2025understanding}, which removes the need for a value function by leveraging multiple responses per prompt. For example, GRPO~\citep{shao2024deepseekmath} computes advantages via group-relative rewards, avoiding critic training while remaining efficient. GSPO~\citep{zheng2025group} improves stability by moving from token- to sequence-level optimization, and RLOO~\citep{ahmadian2024back} uses a leave-one-out baseline for variance reduction. Despite these differences, they all rely on scalar rewards and treat samples independently, ignoring their structural relationships.

\subsection{Reward Models}
Reward models (RMs) are central to RLHF, serving as learned proxies for human preferences and providing scalar rewards for policy optimization~\citep{christiano2017deep}. The standard formulation follows the Bradley--Terry model~\cite{bradley1952rank}, which assigns scalar scores based on pairwise preference data~\cite{bai2022constitutional,ganguli2022red}. Recent work extends this paradigm with more fine-grained supervision, such as process reward models (PRMs) that provide step-level signals~\cite{lightman2023prm}. To address the limitations of raw rewards, a related line of work explores reward shaping techniques that modify or augment reward signals to improve training stability, credit assignment, and robustness to reward hacking~\cite{kwon2023reward,fu2025reward}.

Despite these efforts, most RLHF pipelines ultimately rely on a single scalar reward derived from the final hidden state, compressing rich internal representations into a point estimate~\cite{wang2024secrets}. This overlooks the information encoded in RM representations, which capture semantic and preference-aligned relationships among samples~\cite{yang2024regularizing,wang2024lrhp}. In this work, we instead leverage RM hidden states as reward features to refine advantage estimation.

\begin{figure*}[t]
	\centering
	\includegraphics[width=0.95\textwidth]{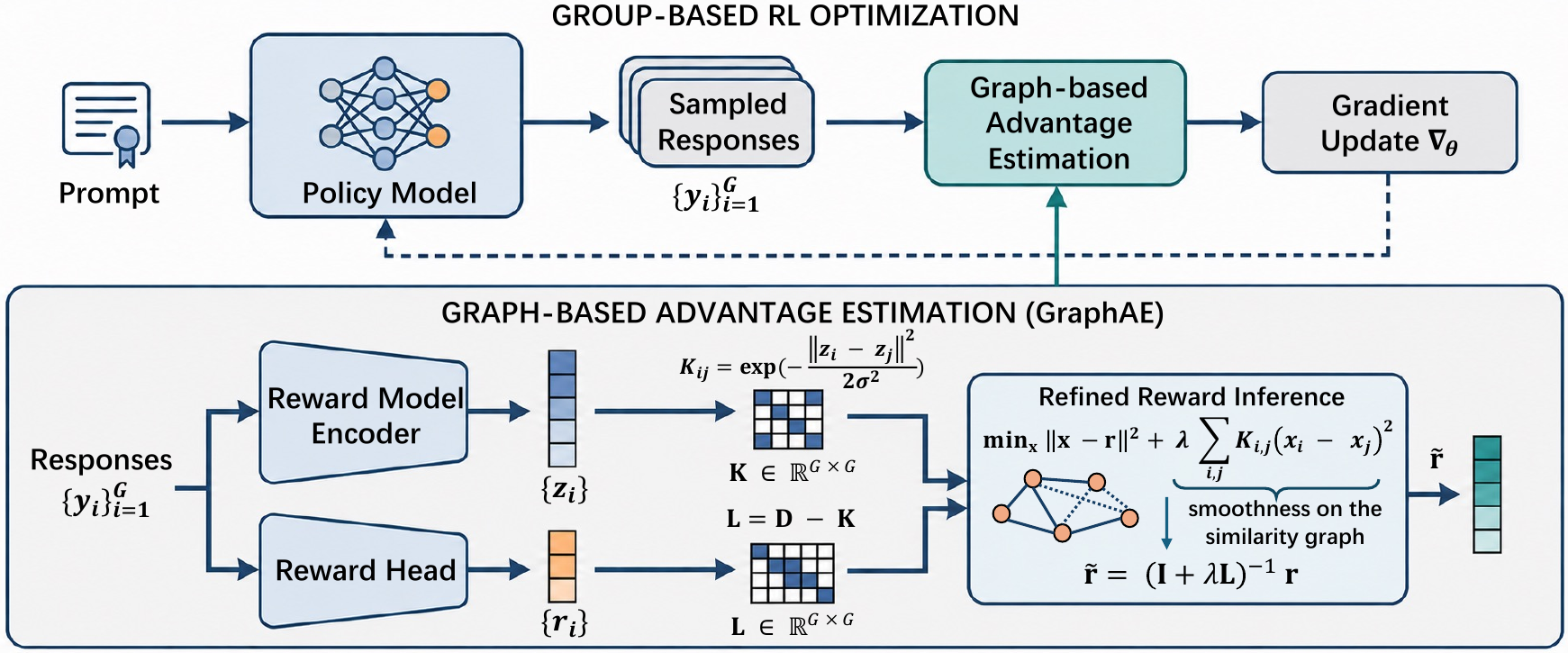}
  	\caption{Overview of GraphAE. GraphAE refines advantages by leveraging RM representation structures, acting as a structured control variate that preserves the reward signal while reducing noise and improving learning efficiency.}
	\label{framework}
    % \vspace{-4mm}
\end{figure*}

%\section{Method}

%\subsection{Preliminary}
\section{Preliminary}
Consider a large language model (LLM) $\pi_\theta$ parameterized by $\theta$, which takes an input sequence $x$ and generates an output sequence $y$ according to the conditional distribution $\pi_\theta(\cdot \mid x)$. Reinforcement learning from human feedback (RLHF) aims to optimize $\pi_\theta$ to maximize expected rewards aligned with human preferences. In practice, such rewards are provided by a reward model (RM) trained from human preference data. Specifically, given a dataset of preference comparisons $\mathcal{D}_{\text{pref}} = \{(x, y_w, y_l)\}$, where $y_w \succ y_l$ indicates that $y_w$ is preferred over $y_l$ for prompt $x$, preferences are commonly modeled using the Bradley--Terry (BT) model~\cite{bradley1952rank} with an underlying ground-truth reward function $r^\star(\cdot, \cdot)$:
\begin{equation}
P(y_w \succ y_l \mid x) = \sigma\big(r^\star(x, y_w) - r^\star(x, y_l)\big),
\end{equation}
where $\sigma(z) = 1 / (1 + e^{-z})$ denotes the sigmoid function. A parametric RM $r(x, y)$ can then be learned via maximum likelihood estimation:
\begin{equation}
\mathcal{L}_{\text{RM}}(r) = \mathbb{E}_{\mathcal{D}_{\text{pref}}} 
\left[-\log \sigma\big(r(x, y_w) - r(x, y_l)\big)\right].
\end{equation}

In RLHF, the RM provides a scalar reward for each response: $r_i = r(x, y_i)$, where $y_i \sim \pi_\theta(\cdot \mid x)$. Based on these rewards, group-based RL algorithms such as GRPO~\cite{shao2024deepseekmath} estimate advantages based on a global baseline:
\begin{equation}
A_i = r_i - \bar{r}, \quad \text{where} \quad \bar{r} = \frac{1}{G} \sum_{j=1}^{G} r_j,
\end{equation}
for a group of $G$ sampled completions. Then RLHF formulates policy optimization as maximizing the expected advantages under the policy while constraining deviation from a reference policy $\pi_{\text{ref}}$:
\begin{equation}
\begin{aligned}
\max_{\theta} \ \mathcal{J}(\theta)
= \ & \mathbb{E}_{x \sim \mathcal{D},\, y \sim \pi_\theta(\cdot \mid x)}
\big[ A (x, y) \big] \\
& - \beta \, \mathbb{E}_{x \sim \mathcal{D}}
\big[ D_{\mathrm{KL}}(\pi_\theta \,\|\, \pi_{\text{ref}}) \big].
\end{aligned}
\end{equation}
where $\beta > 0$ controls the strength of KL regularization. This objective is typically optimized using policy gradient methods~\cite{sutton1999policy}.

% This formulation treats each sample independently and relies solely on scalar rewards, implicitly assuming no structural relationship among the sampled completions. However, this assumption overlooks an important property of modern RMs: beyond scalar outputs, they produce rich internal representations that encode semantic and preference related information. Empirically, these representations capture meaningful relationships between samples, where distances in the representation space correlate with reward differences and preference gaps. This suggests that the RM defines a latent \emph{preference geometry} over the set of sampled completions. 

% Motivated by this observation, we argue that advantage estimation should not treat samples independently, but instead account for their underlying structure. To this end, we reformulate advantage estimation as a \textbf{structured inference problem}: given a set of samples $\{y_i\}_{i=1}^G$, we jointly estimate the advantage vector $\mathbf{A} = [A_1, \dots, A_G]$ over a graph induced by their reward representations, enabling information sharing among similar samples.

%\subsection{Graph Advantage Estimation}
\section{Representation-Aware Advantage Estimation}
We introduce a representation-aware advantage estimation that leverages reward model representations to refine advantage signals via graph regularization. We formalize the estimation objective in \S\ref{sec:formulation} and derive its closed-form solution in \S\ref{sec:solution}. Finally, we show that the refined advantages can be seamlessly integrated into existing group-based RL algorithms in \S\ref{sec:groupl}.

\subsection{Graph-Regularized Advantage Formulation}
\label{sec:formulation}
Existing advantage estimation relies on independent scalar rewards. However, prior work~\citep{zhang2025policy} suggests that reward signals are inherently noisy, as responses with similar quality can still receive noticeably different scores. This issue causes high-variance advantage estimates. To address this issue, we refine the reward signal by introducing two complementary objectives: 
\paragraph{Representation-aware smoothness.}
We encourage similar responses to have consistent reward estimates by enforcing smoothness in the reward model representation space. To this end, we construct a similarity graph over sampled responses. Each node corresponds to a response $y_i$, represented by:
\[
z_i = f_\phi(x, y_i),
\]
where $z_i$ is the hidden representation before the reward head. Edges encode pairwise similarity between responses, with weights defined by a Gaussian kernel:
\begin{equation}
K_{ij} = \exp\left(-\frac{\|z_i - z_j\|^2}{2\sigma^2}\right).
\end{equation}
This yields a fully-connected weighted graph that captures local relationships in the representation space, and induces a smoothness constraint $\sum_{i,j}K_{ij}(x_i - x_j)^2$ that penalizes discrepancies between similar responses.

\paragraph{Fidelity to the original reward.}
To prevent over-smoothing, the refined reward should remain close to the original reward signal produced by the reward model. We therefore impose a fidelity constraint of the form $\|\mathbf{x} - \mathbf{r}\|^2$, which anchors the optimized rewards to the original values.

\vspace{3mm}
Combining the above two objectives, we obtain the refined reward $\tilde{\mathbf r}$ as:
\begin{equation}
\tilde{\mathbf r}
=
\arg\min_{\mathbf x }
\underbrace{\|\mathbf x-\mathbf r\|^2}_{\text{fidelity}}
+
\lambda
\underbrace{\sum_{i,j}K_{ij}(x_i-x_j)^2}_{\text{smoothness}}
\label{eq:objective}
\end{equation}
where $\lambda$ controls the trade-off between fidelity and smoothness.

\subsection{Solution and Properties} 
\label{sec:solution}
The objective in Eq.~\ref{eq:objective} corresponds to a standard graph Laplacian regularization problem. Let $L = D - K$ denote the graph Laplacian, where $D_{ii} = \sum_j K_{ij}$. The optimization problem can be rewritten as:
\begin{equation}
\tilde{\mathbf r} = \arg\min_{\mathbf x} 
\|\mathbf x - \mathbf r\|^2 + \lambda \mathbf x^\top L \mathbf x.
\end{equation}
Taking derivatives with respect to $\mathbf x$ and setting to zero gives:
\begin{equation}
2(\mathbf x - \mathbf r) + 2\lambda L \mathbf x = 0,
\end{equation}
which yields the linear system:
\begin{equation}
(\mathbf I + \lambda L)\mathbf x = \mathbf r.
\end{equation}
Since $\mathbf I + \lambda L \in \mathbb{R}^{G \times G}$ is symmetric positive definite, the solution is unique and can be computed efficiently. In RLHF, the group size $G$ is typically small (e.g., $G \leq 32$), making direct solvers both efficient and numerically stable. The computational cost is $O(G^3)$ and negligible in practice. Therefore, the optimal solution admits a closed form:
\begin{equation}
\tilde{\mathbf r} = (\mathbf I + \lambda L)^{-1} \mathbf r.
\label{eq:closed_form}
\end{equation}
This solution can be interpreted as graph-based smoothing of the reward signal, where each entry aggregates information from similar responses while remaining anchored to its original value.

\subsection{Compatibility with Group-based RL}
\label{sec:groupl}
Our method can be seamlessly integrated into a wide range of group-based RL algorithms. Specifically, we replace the original reward $r_i$ with the refined reward $\tilde{r}_i$ in the advantage estimator. 
For example, in GRPO ~\cite{shao2024deepseekmath} :
\begin{equation}
A_i^{\text{GRPO}}:
\quad
\frac{r_i - \mu_r}{\sigma_r + \epsilon}
\;\rightarrow\;
\frac{\tilde{r}_i - \mu_{\tilde{r}}}{\sigma_{\tilde{r}} + \epsilon}.
\end{equation}
This modification only affects the advantage estimation step, while leaving the policy optimization procedure unchanged.

\section{Experimental Setup}
%\subsection{Models} 
%We adopt an open-source reward model Skywork-Reward-V2-Llama-3.1-8B~\cite{liu2025skywork} without training from scratch, which is trained on large-scale preference data and has state-of-the-art performance across multiple benchmarks. This design allows us to isolate the contribution of representation advantage estimation without confounding factors from training. During reward computation, besides the scalar reward, we additionally extract the representation of the last token from the final layer, which is used for our graph-based advantage estimation. For the policy model, we conduct RL training on two widely used instruction-tuned LLMs: Qwen2.5-7B-Instruct~\cite{yang2025qwen3} and Llama-3-8B-Instruct~\cite{grattafiori2024llama}.

%For policy training, we use the prompts from Ultrafeedback~\cite{cui2024ultrafeedback}, which consists of over 60k prompts covering various domains. Then 
\subsection{Datasets and Evaluation} 
We evaluate the performance of GraphAE on three widely adopted RLHF benchmarks: Arena-Hard-v0.1~\cite{li2024crowdsourced}, AlpacaEval 2.0~\cite{dubois2024length}, and MT-Bench~\cite{zheng2023judging}. These benchmarks require the model to generate answers to a wide range of open-ended questions.

We follow the standard evaluation protocol of each benchmark. For Arena-Hard-v0.1, we report the win rate (WR) against GPT-4-0314 with the 95\% confidence interval (CI), using GPT-4.1 as the judge. For AlpacaEval 2.0, we report both WR and length-controlled win rate (LC WR) against GPT-4-Turbo. For MT-Bench, we report two turns and average scores using GPT-4-Turbo as the judge. More details are provided in the Appendix \ref{datasets}.

% \subsection{Baselines} We first use the Qwen2.5-7B-Instruct and Llama-3-8B-Instruct models, which have not undergone RL training, as performance references. To evaluate the effectiveness and generality of GraphAE, we compare with a diverse set of RL algorithms, including PPO~\cite{schulman2017proximal}, and group-based RL methods such as GRPO~\cite{shao2024deepseekmath}, GSPO~\cite{zheng2025group}, and RLOO~\citep{ahmadian2024back}. For all RL methods, we use identical training configurations across all algorithms to ensure a fair comparison. Under this unified setting, we only replace the advantage estimator with our proposed GraphAE. This design isolates the impact of our method from other factors and allows a direct evaluation of its contribution. 

\subsection{Baselines}
GraphAE is designed as a plug-in advantage estimator for group-based RLHF. We apply it to three representative group-based RL algorithms, including GRPO~\cite{shao2024deepseekmath}, GSPO~\cite{zheng2025group}, and RLOO~\citep{ahmadian2024back}. For each method, we replace the original scalar reward based advantage estimator with GraphAE while keeping all other components unchanged.

We compare GraphAE against RL methods that use scalar rewards for advantage estimation. To avoid the effect of reward regularization, we further include reward smoothing baselines of the form $\tilde r_i = (1-\alpha) r_i + \alpha \bar r$, where $\bar r$ denotes the mean reward within each rollout group and $\alpha$ controls the smoothing strength.\footnote{The $\alpha$ is set to 0.3 across all RL settings based on experiments in Appendix \ref{app:alpha}} For a more comprehensive evaluation, we also include classical baselines such as PPO~\cite{schulman2017proximal} and the original instruction-tuned models without RL training.

\subsection{Implementation Details}
\paragraph{Reward Model.} We adopt a pre-trained reward model Skywork-Reward-V2-Llama-3.1-8B~\cite{liu2025skywork} given its superior performance across multiple benchmarks. The pre-trained model allows us to isolate the contribution of representation advantage estimation without confounding factors from training. In GraphAE, we extract the representation of the last token from the final layer for graph-based advantage estimation.
\paragraph{Policy Training.} We conduct RL training on two widely used instruction-tuned LLMs: Qwen2.5-7B-Instruct~\cite{yang2025qwen3} and Llama-3-8B-Instruct~\cite{grattafiori2024llama}. Specifically, we use a learning rate of 2e-6 and set the maximum generated sequence length to 1024. We employ a batch size of 1024 and a mini-batch size of 8, with 1 epoch and 128 gradient accumulation steps. The sampling temperature is set to 0.7 and the KL coefficient is set to 0.01. We roll out 8 responses for each prompt. We set the smoothing coefficient $\lambda$ to 1 for Qwen2.5-7B-Instruct and 0.5 for Llama-3-8B-Instruct. For the Gaussian kernel bandwidth $\sigma$, we adaptively set it to the median of all non-zero pairwise distances within each group, allowing it to adjust to different group scales.
%\textcolor{blue}{ For the reward smoothing baseline, we use a fixed smoothing coefficient $\alpha=0.3$ across all settings. This provides a simple uniform regularization baseline without introducing additional tuning complexity.}

\begin{table*}[t]
\centering
\small
\setlength{\tabcolsep}{3.2pt}
\begin{tabular}{l cc cc ccc | cc cc ccc}
\toprule

\multirow{2}{*}{\textbf{Method}}
& \multicolumn{7}{c|}{\textbf{Qwen2.5-7B-Instruct}}
& \multicolumn{7}{c}{\textbf{Llama-3-8B-Instruct}} \\

\cmidrule(lr){2-8}
\cmidrule(lr){9-15}

& \multicolumn{2}{c}{\textbf{Arena-Hard}}
& \multicolumn{2}{c}{\textbf{AlpacaEval}}
& \multicolumn{3}{c|}{\textbf{MT-Bench}}

& \multicolumn{2}{c}{\textbf{Arena-Hard}}
& \multicolumn{2}{c}{\textbf{AlpacaEval}}
& \multicolumn{3}{c}{\textbf{MT-Bench}} \\

\cmidrule(lr){2-3}
\cmidrule(lr){4-5}
\cmidrule(lr){6-8}
\cmidrule(lr){9-10}
\cmidrule(lr){11-12}
\cmidrule(lr){13-15}

& WR & CI
& LC WR & WR
& T1 & T2 & Avg

& WR & CI
& LC WR & WR
& T1 & T2 & Avg \\

\midrule

w/o RL
& 45.7 & -2.1/+1.6
& 33.74 & 33.41
& 8.23 & 6.96 & 7.59

& 19.0 & -1.7/+1.5
& 22.09 & 22.82
& 7.19 & 6.05 & 6.62 \\

PPO
& 52.9 & -1.7/+1.9
& 35.60 & 35.10
& 8.30 & 7.15 & 7.73

& 33.2 & -1.7/+1.9
& 34.60 & 34.20
& 7.35 & 6.95 & 7.15 \\

\midrule

GRPO
& 51.5 & -1.9/+1.7
& 36.26 & 35.94
& \textbf{8.66} & 7.01 & 7.84

& 31.1 & -1.8/+1.9
& 34.85 & 35.53
& 7.64 & 6.63 & 7.13 \\

GRPO + Smooth
& 51.0 & -2.0/+1.7
& 36.40 & 36.12
& 8.61 & 7.02 & 7.82

& 30.8 & -1.9/+1.9
& 34.91 & 35.62
& 7.60 & 6.62 & 7.11 \\

\rowcolor{lightgreen}
GRPO + GraphAE
& \textbf{57.8} & -1.9/+2.2
& \textbf{42.85}
& \textbf{44.21}
& 8.60 & \textbf{7.10} & \textbf{7.85}

& \textbf{33.9} & -1.6/+1.8
& \textbf{37.56}
& \textbf{36.03}
& \textbf{7.68} & \textbf{6.74} & \textbf{7.21} \\

\midrule

GSPO
& 55.5 & -2.4/+2.1
& 35.76 & 36.23
& 8.50 & 7.23 & 7.86

& 31.6 & -1.5/+1.8
& 36.47 & 36.98
& 7.49 & 6.53 & 7.01 \\

GSPO + Smooth
& 54.4 & -2.1/+2.2
& 35.93 & 36.40
& 8.48 & 7.21 & 7.85

& 30.4 & -1.7/+1.9
& 36.51 & 37.01
& 7.46 & 6.52 & 6.99 \\

\rowcolor{lightgreen}
GSPO + GraphAE
& \textbf{59.5} & -1.9/+1.8
& \textbf{40.63}
& \textbf{42.05}
& \textbf{8.60} & \textbf{7.43} & \textbf{8.01}

& \textbf{34.2} & -2.1/+1.8
& \textbf{37.85}
& \textbf{37.92}
& \textbf{7.64} & \textbf{6.83} & \textbf{7.23} \\

\midrule

RLOO
& 50.8 & -1.9/+2.2
& 35.90 & 35.20
& 8.55 & 7.05 & 7.80

& 30.5 & -1.8/+1.7
& 34.23 & 35.81
& 7.55 & 6.60 & 7.05 \\

RLOO + Smooth
& 50.8 & -1.7/+2.0
& 35.81 & 35.42
& 8.50 & 7.04 & 7.77

& 30.1 & -1.6/+1.8
& 34.31 & 35.94
& 7.51 & 6.58 & 7.05 \\

\rowcolor{lightgreen}
RLOO + GraphAE
& \textbf{55.2} & -1.8/+2.1
& \textbf{39.50}
& \textbf{41.10}
& \textbf{8.60} & \textbf{7.20} & \textbf{7.98}

& \textbf{32.7} & -1.9/+1.7
& \textbf{36.13}
& \textbf{36.81}
& \textbf{7.62} & \textbf{6.75} & \textbf{7.16} \\

\bottomrule
\end{tabular}

\caption{
Main results on Qwen2.5-7B-Instruct and Llama-3-8B-Instruct across Arena-Hard-v0.1, AlpacaEval 2.0, and MT-Bench. Smooth denotes the uniform reward smoothing baseline with $\alpha=0.3$.
}

\label{tab:main_results}
\vspace{-2mm}
\end{table*}

\section{Results and Analysis} 
\subsection{Main Results}

% Table~\ref{tab:main_results} summarizes the main results in three evaluation benchmarks.
% Compared to the base models without RL training, all RL methods achieve substantial improvements, confirming the effectiveness of online RL for alignment.
% Among these methods, PPO provides moderate gains but requires a separate value network, increasing training complexity. In contrast, group-based methods achieve comparable or better performance without a critic model. When combined with GraphAE, all three group-based algorithms show further improvements across both policy backbones and all evaluation benchmarks. Unlike GraphAE, uniform reward smoothing fails to improve RLHF performance, which shows that the improvements do not come merely from smoothing or regularizing rewards, but from exploiting the structured preference information encoded in RM representations.
% These results show that GraphAE is a general-purpose module that benefits different group-based RL algorithms without modifying their optimization objectives. While GraphAE only changes how advantages are estimated, it does not modify the reward model or the training objective. The consistent gains across all algorithms and backbones suggest that scalar rewards alone miss fine-grained preference signals that RM hidden states retain.

\paragraph{Overall Results.}
Table~\ref{tab:main_results} summarizes the main results across three evaluation benchmarks. Compared with the original group-based RL algorithms, GraphAE consistently improves GRPO, GSPO, and RLOO across both model backbones and all evaluation benchmarks. This demonstrates the effectiveness and generalization ability of GraphAE, compared to original scalar based advantage estimation methods. We also observe that the uniform reward smoothing baseline (+Smooth) yields only marginal or inconsistent improvements, suggesting that the gains of GraphAE are not attributable to reward regularization, but rather to better exploitation of preference information encoded in RM representations. Moreover, all RL methods substantially outperform the w/o RL baselines, while PPO achieves only moderate gains despite requiring an additional value network.

%\textcolor{blue}{The largest gains appear on Arena-Hard and AlpacaEval, with up to +6.3 and +8.27 over GRPO. The improvements on MT-Bench are smaller. This difference is expected: Arena-Hard and AlpacaEval contain responses with subtle quality differences, making scalar rewards noisy and advantage estimation inaccurate. GraphAE reduces this noise by propagating rewards over the graph. MT-Bench, by contrast, has a narrower score range among strong models, leaving less room for improvement. The consistent gains of GraphAE across all three benchmarks, larger in noisy settings and smaller in saturated ones, suggest that RM hidden states encode useful structural information beyond scalar rewards. }

GraphAE achieves its largest gains on Arena-Hard and AlpacaEval, improving over GRPO by up to +6.3 and +8.27 points, respectively, while the gains on MT-Bench are comparatively smaller. This pattern is consistent with the characteristics of the benchmarks. In Arena-Hard and AlpacaEval, the quality differences between responses are often subtle, making scalar rewards more susceptible to noise and leading to less reliable advantage estimates. GraphAE alleviates this issue by leveraging representation aware advantage estimation, which provide richer and more stable signals. In contrast, MT-Bench exhibits a narrower score range among strong models, which limits the headroom for further improvement.

%\textcolor{blue}{A further pattern is that GraphAE brings larger gains on Qwen than on Llama, while remaining consistently effective across both backbones. We attribute this to stronger models producing more informative within-group structure in RM representation space, which provides better signals for graph-based refinement.}

%\xiyan{{\paragraph{Feature Ablation.} In this section, we explore whether the improvements of GraphAE come from the preference relevant information encoded in RM hidden states, rather than merely from introducing a graph structure. To examine this, }

\paragraph{Feature Ablation.} In this section, we explore whether the improvements of GraphAE come from the preference relevant information encoded in RM hidden states, rather than merely from introducing a graph structure. To examine this, we conduct a feature ablation by keeping the underlying Qwen + GRPO optimization pipeline unchanged and varying only the feature source used for graph construction. We compare GraphAE with three controlled variants: i) \emph{Scalar-Graph}, which constructs graph edges from scalar reward differences; ii) \emph{Policy-Graph}, which uses the last-token hidden states of the policy model; and iii) \emph{Shuffled RM-Graph}, which randomly permutes the correspondence between responses and RM features.

\begin{figure}[t]
    \centering
    \includegraphics[width=0.99\linewidth]{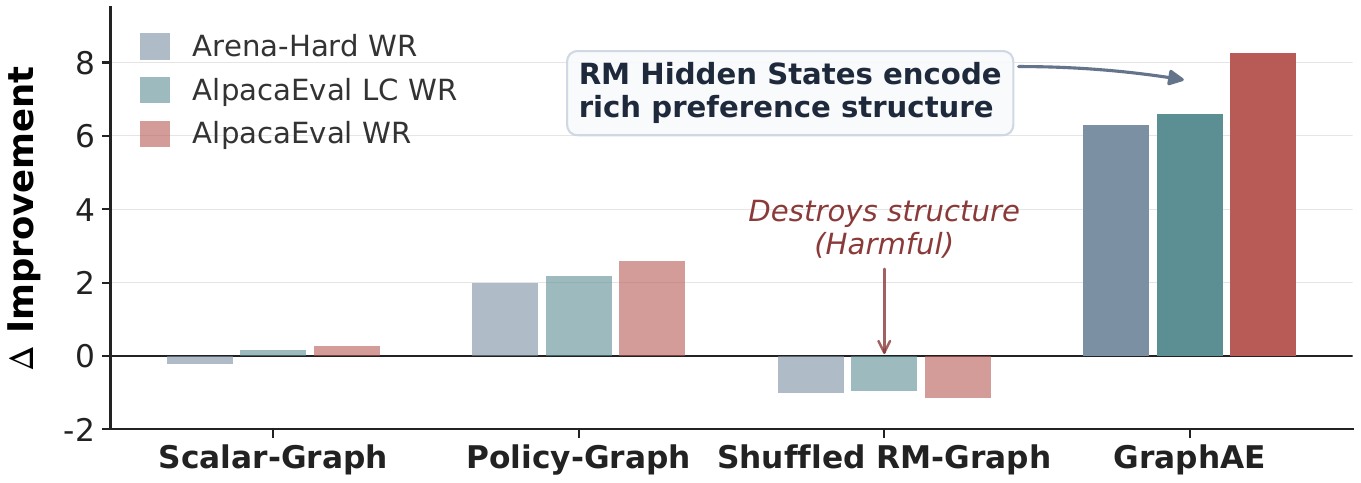}
    \caption{Ablation of graph construction strategies.}
    \label{fig:reward_feature_ablation}
    \vspace{-4mm}
\end{figure}

As shown in Figure~\ref{fig:reward_feature_ablation}, \emph{Scalar-Graph} performs similarly to vanilla GRPO, with slight drops on Arena-Hard and only marginal gains on AlpacaEval. This suggests that graph construction based solely on scalar rewards provides little additional information beyond the reward signal already used by GRPO. \emph{Policy-Graph} achieves moderate improvements, indicating that policy-model representations capture some local semantic structure among responses. However, the remaining gap to GraphAE suggests that semantic similarity alone is insufficient for constructing an effective preference graph. In contrast, \emph{Shuffled RM-Graph} degrades performance below GRPO across all metrics, demonstrating that an incorrect neighborhood structure can actively harm optimization. Overall, these results show that the gains of GraphAE primarily stem from the preference-relevant local geometry encoded in RM hidden states, rather than from graph connectivity, feature dimensionality, or the use of a graph structure alone.

%\textcolor{blue}{\paragraph{Feature Ablation.} We keep the Qwen + GRPO training setup fixed and vary only the feature source used to construct the graph. We compare the full method against three controlled variants: \emph{Reward-Graph}, which constructs edges directly from scalar reward differences; \emph{Policy-Graph}, which uses policy model last token hidden states; and \emph{Shuffled RM-Graph}, which randomly permutes the correspondence between responses and RM features. As shown in Figure~\ref{fig:reward_feature_ablation}, \emph{Reward-Graph} performs similarly to vanilla GRPO, with slight drops on Arena-Hard and only marginal gains on AlpacaEval. This is expected, as constructing the graph from scalar rewards introduces no additional information. \emph{Policy-Graph} achieves moderate improvements, indicating that generic semantic features capture some useful local structure. However, the gap to GraphAE remains substantial, suggesting that semantic similarity alone is insufficient. Crucially, the shuffled variant degrades performance below GRPO across all metrics, showing that incorrect neighborhood structure is harmful. These results demonstrate that GraphAE’s gains stem from the correct local preference geometry induced by RM features, rather than graph connectivity or feature dimensionality alone.}

% \textcolor{blue}{A further pattern is that GraphAE brings larger gains on Qwen than on Llama, though improvements remain positive in all settings. We attribute this to the quality of within-group structure in RM representation space. Qwen, as a stronger model, produces responses with clearer quality distinctions. These distinctions are well captured by RM hidden states, forming informative pairwise relations for graph propagation. Llama, being a weaker model, produces responses with less differentiated quality, resulting in a less structured graph that offers fewer opportunities for refinement. Still, GraphAE yields positive gains even in this case, showing that the method is robust across different model capacities.}

% \subsection{Analysis and Generalization}

% We further analyze why GraphAE improves RLHF training from three perspectives. First, we study whether GraphAE improves training stability and credit assignment by denoising noisy within-group reward signals. Second, we investigate whether the gain comes from the specific preference geometry encoded in RM representations, rather than graph smoothing or regularization alone. Finally, we examine whether this representation structure generalizes across different reward models.

\subsection{Mechanism Analysis}
\begin{figure}[t]
    \centering
    \includegraphics[width=0.9\linewidth]{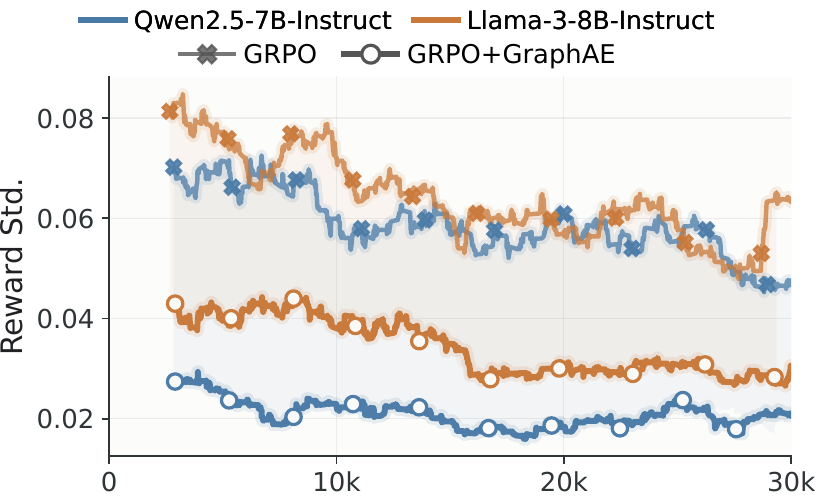}
    \caption{Same budget training dynamics of group reward std. over the first 30k training steps on GRPO.}
    \label{fig:reward_std_grpo}
    \vspace{-4mm}
\end{figure}

To better understand why GraphAE improves performance, we conduct a series of analyses on its underlying behavior:

\paragraph{Variance Reduction.} We compare training steps of GRPO under the same training settings. For each training step, we compute the standard deviation of rewards within each rollout group and track its moving average throughout training. A lower value means that rewards within the same rollout group are less variable. As illustrated in Figure~\ref{fig:reward_std_grpo}, we observe that GraphAE reduces the within-group reward standard deviation, compared with the original scalar reward based optimization. This suggests that GraphAE provides smoother and less noisy rewards, enhancing the training stability.

%\paragraph{Mechanism Analysis.}
%\textcolor{blue}{We first test whether the gains of GraphAE arise from improving noisy within-group reward signals. To answer this, we analyze the mechanism of GraphAE from three complementary perspectives: we first study whether GraphAE stabilizes within-group rewards during training, then examine whether this effect preserves meaningful preference structure rather than simply over-smoothing rewards, and finally analyze whether the benefit becomes stronger in the high-dispersion groups where scalar reward based credit assignment is most unreliable.}

%\textcolor{blue}{We begin by studying training stability. Figure~\ref{fig:reward_std_all_methods} compares the first 30k training steps of different RL methods under the same training budget. At each step, we compute the standard deviation of rewards within each rollout group and track its moving average over training. Lower values indicate less dispersed within-group reward signals and therefore reduced reward variability during training. The results show that GraphAE consistently reduces group reward dispersion across all backbones, suggesting more stable training dynamics.}

\begin{figure}[t]
    \centering
    \includegraphics[width=\linewidth]{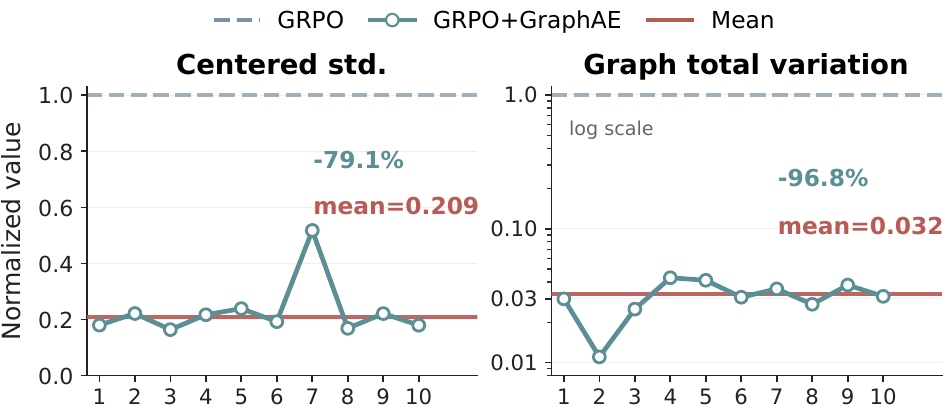}
    \caption{Denoising effect of GraphAE under GRPO.}
    \label{fig:denoising_metrics_qwen_grpo}
    \vspace{-2mm}
\end{figure}

\paragraph{Denoising Without Losing Preference Structure.} After observing more stable within-group rewards during training, we further ask whether GraphAE preserves the underlying preference structure while reducing reward noise. We analyze 120 rollout groups with RM features. For each group, we compare the original reward signal with the GraphAE signal from Eq.~\ref{eq:closed_form} using i) centered-signal standard deviation, $\mathrm{Std}(s) = \left( \frac{1}{G} \sum_{i=1}^{G} \left(s_i - \bar{s}\right)^2 \right)^{1/2}$, where $\bar{s}=\frac{1}{G}\sum_{i=1}^{G}s_i$; and ii) graph total variation, $\mathrm{GTV}(s) = \frac{1}{G^2} \sum_{i=1}^{G} \sum_{j=1}^{G} K_{ij} \left(s_i-s_j\right)^2$, where $K_{ij}$ denotes the RM hidden-state graph edge weight. We report both metrics after normalization by the corresponding value of the original scalar reward signal.
%Both metrics are normalized so that the raw reward signal equals 1.0. 
Figure~\ref{fig:denoising_metrics_qwen_grpo} shows ten randomly sampled results, and the average value across 120 groups.

%For a rollout group with reward signal $s \in \mathbb{R}^{G}$, we compute the centered-signal standard deviation
%$$
%    \mathrm{Std}(s) = \left( \frac{1}{G} \sum_{i=1}^{G} \left(s_i - \bar{s}\right)^2 \right)^{1/2}, \bar{s}=\frac{1}{G}\sum_{i=1}^{G}s_i
%$$
%and the graph total variation
%$$
%    \mathrm{GTV}(s) = \frac{1}{G^2} \sum_{i=1}^{G} \sum_{j=1}^{G} K_{ij} \left(s_i-s_j\right)^2
%$$
%where $K_{ij}$ denotes the RM hidden-state graph edge weight. We report both metrics after normalization by the corresponding value of the original scalar reward signal.

GraphAE substantially reduces both quantities, showing that it effectively suppresses local reward noise. However, the denoised signal still remains highly aligned with the original reward structure: the average pairwise sign-flip rate is only 6.25\%, and the cosine alignment between the original rewards and GraphAE remains 0.905. These results suggest that GraphAE reduces noisy local fluctuations without collapsing the preference geometry.

% \begin{figure}[t]
%     \centering
%     \includegraphics[width=\linewidth]{figures/fig_mechanism_denoising_bins_qwen_gspo.pdf}
%     \caption{Mechanism analysis on GSPO with GraphAE.}

%     \label{fig:denoising_and_bins}
%     \vspace{-4mm}
% \end{figure}

% \begin{figure}[t]
%     \centering
%     \includegraphics[width=\linewidth]{figures/fig_mechanism_gain_by_dispersion_qwen_grpo.pdf}
%     \caption{Gains under different reward variance levels.}
%     \label{fig:dispersion_qwen_grpo}
%     \vspace{-4mm}
% \end{figure}

\begin{table}[t]
\centering
\small
\setlength{\tabcolsep}{5.2pt}
\renewcommand{\arraystretch}{1.12}
\begin{tabular}{lcc}
\toprule
\textbf{Baseline Reward std.} & \textbf{Reward Gain} & \textbf{Positive Rate} \\ \midrule $<$2.5 & -1.4 & 38.1\% \\ 2.5$\sim$5.0 & -1.0 & 45.4\% \\ 5.0$\sim$7.5 & +0.6 & 54.0\% \\ $>$7.5 & \textbf{+10.4} & \textbf{70.8\%} \\
\bottomrule
\end{tabular}
\caption{Reward gains under different variance levels. Positive rate is the fraction of groups in each bucket where GraphAE increases the group mean reward.}
\label{tab:gain_by_dispersion_qwen_grpo}
\vspace{-4mm}
\end{table}

%\textcolor{blue}{We next examine whether GraphAE simply smooths rewards or preserves meaningful preference structure during denoising. The left panel of Figure~\ref{fig:denoising_and_bins} analyzes 128 rollout groups with RM features. For each group, we compare the original reward signal with the GraphAE signal from Eq.~\ref{eq:closed_form} using centered-signal standard deviation and graph total variation. Both metrics are normalized so that the raw reward signal equals 1.0. GraphAE substantially reduces both quantities, showing that it effectively suppresses local reward noise. However, the denoised signal still remains highly aligned with the original reward structure: the average pairwise sign-flip rate is only 5.3\%, and the cosine alignment between the original rewards and GraphAE remains 0.889. These results suggest that GraphAE reduces noisy local fluctuations without collapsing the underlying preference geometry.}

\paragraph{Stronger Benefits for High-Variance Group.} Since GraphAE reduces local reward fluctuations without destroying preference structure, we next examine where this denoising effect brings the largest benefit. We bucket matched rollout groups by their baseline reward standard deviation under vanilla GRPO, and measure the change in group mean reward after applying GRPO + GraphAE on the same prompts. Table~\ref{tab:gain_by_dispersion_qwen_grpo} shows that GraphAE's gain is concentrated in high-variance groups. Low-variance groups receive limited or slightly negative gains, while the highest-variance bucket shows a clear positive improvement. This suggests that GraphAE is most useful when scalar rewards are noisy and credit assignment is unreliable, rather than uniformly smoothing all groups.

\subsection{Robustness and Efficiency}
\paragraph{Reward Model Generalization.}

We next test whether the preference geometry exploited by GraphAE is specific to one RM. To this end, we conduct experiments using two additional RMs, \textit{Skywork-Reward-V2-Qwen3-8B}~\cite{liu2025skywork} and \textit{Eurus-RM-7B}~\cite{yuan2024advancing} and further report results across multiple group-based RL algorithms, including GRPO, GSPO, and RLOO. In addition to downstream performance, we measure the correlation between RM representation distances and reward differences, $\rho(d_h, |\Delta r|)$, as a proxy for how strongly preference structure is encoded in the representation space.

As shown in Table~\ref{tab:rm_generalization}, hidden state geometry consistently correlates with reward gaps across all RMs, indicating that preference structure is broadly encoded in RM representations. Importantly, the strength of this correlation closely aligns with downstream gains: models with better representations yield larger improvements under GraphAE, while weaker RMs lead to reduced but still positive gains. We further observe that GraphAE improves performance consistently across GRPO, GSPO, and RLOO, suggesting that the benefit is not tied to a specific optimization algorithm. Combined with the consistent trends across different RMs, these results indicate that GraphAE exploits a general and model-agnostic structural property of RM representations rather than relying on particular RMs or RL formulations.

\begin{table}[t]
\centering
\scriptsize
\setlength{\tabcolsep}{3pt}
\begin{tabular}{l c c c c c}
\toprule
\textbf{RM}
& $\rho(d_h, |\Delta r|)$
& \textbf{GRPO}
& \textbf{GSPO}
& \textbf{RLOO} \\
\midrule

Llama-3.1-8B
& 0.873
& +6.3 / +8.27
& +4.0 / +5.82
& +4.4 / +7.50 \\

Qwen3-8B
& 0.861
& +5.8 / +7.52
& +4.2 / +6.05
& +4.1 / +6.88 \\

Eurus-RM-7B
& 0.712
& +2.9 / +4.73
& +2.1 / +3.66
& +2.0 / +3.90 \\

\bottomrule
\end{tabular}
\caption{Reward model generalization across different RMs and RL algorithms.}
\label{tab:rm_generalization}
\vspace{-2mm}
\end{table}

\begin{figure}[t]
    \centering
    \includegraphics[width=\columnwidth]{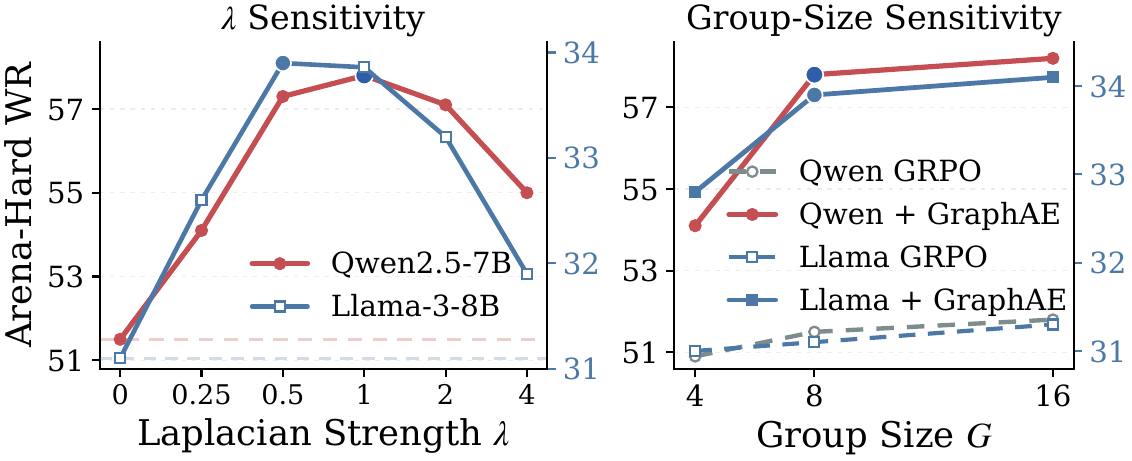}
    \caption{Robustness to the main hyperparameters of GraphAE on Arena-Hard under GRPO.}
    \label{fig:robustness_lambda_group}
    \vspace{-4mm}
\end{figure}

\begin{table}[t]
\centering
\small
\setlength{\tabcolsep}{5pt}
\begin{tabular}{c  c  c}
\toprule
\textbf{Group Size} & \textbf{Compute Latency} & \textbf{Scratch Memory} \\
\midrule
$G=8$  & 0.80 ms & 1.34 KB \\
$G=16$ & 3.29 ms & 5.19 KB \\
$G=32$ & 15.27 ms & 20.38 KB \\
\bottomrule
\end{tabular}
\caption{Compute overhead of the GraphAE.}
\label{tab:efficiency_overhead}
\vspace{-2mm}
\end{table}

\begin{figure}[t]
    \centering
    \includegraphics[width=\columnwidth]{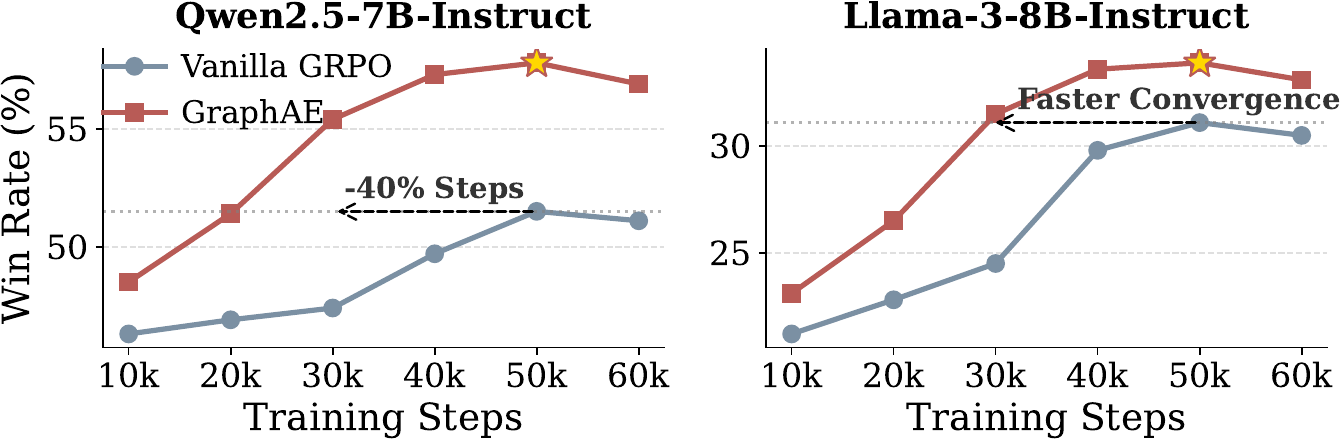}
    \caption{Checkpoint sample efficiency analysis.}
    \label{fig:checkpoint_efficiency}
    \vspace{-4mm}
\end{figure}

\paragraph{Robustness to Hyperparameters.} We evaluate the robustness of GraphAE to its main hyperparameters on Arena-Hard under GRPO. Figure~\ref{fig:robustness_lambda_group} varies the Laplacian strength $\lambda$, and varies the rollout group size $G$ for two backbones. First, GraphAE is not sensitive to a narrow choice of $\lambda$. Both backbones improve substantially once $\lambda>0$ and remain strong over a broad range. The best value is attained near $\lambda=1$ for Qwen and near $\lambda=0.5$ for Llama, but performance around these optima is very similar, indicating that GraphAE does not rely on a sharp sweet spot. Second, GraphAE is also robust to group size. Larger groups bring slightly stronger improvements, where richer local neighborhoods make graph-based credit assignment more effective. These results suggest that GraphAE is stable across reasonable hyperparameter choices rather than relying on fragile tuning.

\paragraph{Overhead.} Table~\ref{tab:efficiency_overhead} shows the estimator overhead. Computing the GraphAE on 4096-dimensional rollout embeddings yields 0.80 ms latency per group at $G=8$, and 15.27 ms at $G=32$, while the scratch memory stays below 21 KB per group. Since the estimator only solves a small $G \times G$ linear system after reward computation, its cost is negligible relative to policy and RM forward passes. Taken together, these results show that GraphAE is both more sample efficient and lightweight in practice.

\paragraph{Checkpoint Efficiency.} 
To investigate whether GraphAE accelerates policy convergence, we save checkpoints every 10k training steps and evaluate them on Arena-Hard. As shown in Figure~\ref{fig:checkpoint_efficiency}, GraphAE improves alignment quality throughout the entire training trajectory: it robustly outperforms vanilla GRPO at every evaluated checkpoint. For Qwen, GraphAE reaches the final GRPO peak score of 51.5 at merely 30k steps, whereas vanilla GRPO requires 50k steps to reach this level, corresponding to a 40\% reduction in optimization steps. This accelerated convergence pattern generalizes remarkably to Llama, where GraphAE attains the GRPO peak in fewer than 30k steps and eventually climbs to an unprecedented 33.9. Moreover, both models show that our method improves both final performance and convergence speed under the same compute budget.

% \begin{table}[t]
% \centering
% \small
% \setlength{\tabcolsep}{4.5pt}
% \begin{tabular}{c c c c c c c}
% \toprule
% \textbf{Method} & \textbf{10k} & \textbf{20k} & \textbf{30k} & \textbf{40k} & \textbf{50k} & \textbf{60k} \\
% \midrule
% GRPO & 46.3 & 46.9 & 47.4 & 49.7 & 51.5 & 51.1 \\
% + GraphAE & 48.5 & 51.4 & 55.4 & 57.3 & \textbf{57.8} & 56.9 \\
% $\Delta$ & +2.2 & +4.5 & +8.0 & +7.6 & +6.3 & +5.8 \\
% \bottomrule
% \end{tabular}
% \caption{Checkpoint sample efficiency analysis.}
% \label{tab:checkpoint_sample_efficiency}
% \end{table}

\section{Conclusion}
We propose to leverage reward model hidden states as graph signals to compute advantages. By propagating rewards over graphs, our method GraphAE captures local correlations, reduces variance, and generalizes group-based RL algorithms. It bridges the gap between scalar reward estimation and structured inference, offering a more information rich, stable, and effective RLHF paradigm.

\clearpage

\newpage

\section*{Limitations}
GraphAE relies on the quality of the reward model representations to construct the similarity graph. When the RM hidden states contain weak or noisy preference structure, the effectiveness of graph-based propagation may be reduced. In addition, GraphAE introduces a small amount of extra computation for graph construction and matrix solving within each rollout group. Although this overhead is lightweight in our experiments, scaling to substantially larger group sizes may require more efficient approximations in future work.

\bibliography{custom}

\clearpage
\appendix

\begin{table*}[ht]
\centering
\small
\setlength{\tabcolsep}{2pt}
\begin{tabular}{lcccccc}
\toprule
Method & Arena-Hard & AlpacaEval LC & AlpacaEval WR & MT-Bench T1 &  MT-Bench T2 &  MT-Bench Avg \\
\midrule
Qwen2.5-7B-Instruct & 45.7 & 33.74 & 33.41 & 8.23 & 6.96 & 7.59 \\
GRPO & 51.5 & 36.26 & 35.94 & \textbf{8.66} & 7.01 & 7.84 \\
\midrule
Reward Smooth ($\alpha=0.1$) & 52.3 & 36.11 & 35.72 & 8.61 & 7.00 & 7.81 \\
Reward Smooth ($\alpha=0.3$) & 51.0 & 36.40 & 36.12 & 8.61 & 7.02 & 7.82\\
Reward Smooth ($\alpha=0.5$) & 49.9 & 35.12 & 34.67 & 8.52 & 6.95 & 7.74 \\
Reward Smooth ($\alpha=0.7$) & 47.7 & 34.51 & 33.98 & 8.40 & 6.88 & 7.64 \\
\midrule
GraphAE & \textbf{57.8} & \textbf{42.85} & \textbf{44.21} & 8.60 & \textbf{7.10} & \textbf{7.85} \\
\bottomrule
\end{tabular}
\caption{Comparison between GraphAE and generic reward smoothing with different coefficient $\alpha$.}
\label{tab:smoothing_ablation}
\end{table*}

\newcolumntype{Y}{>{\raggedright\arraybackslash}X}

\begin{table*}[t]
\centering
\scriptsize
\setlength{\tabcolsep}{4pt}
\renewcommand{\arraystretch}{1.18}
\begin{tabularx}{\textwidth}{p{0.13\textwidth} Y}
\toprule
\multicolumn{2}{l}{\textbf{Qualitative examples of GraphAE reward correction under Qwen2.5-7B-Instruct + GRPO}} \\
\midrule

\multicolumn{2}{l}{\textbf{Case 1: Over-rewarded scientific claim is pulled down}} \\
\textbf{Prompt} &
Why do we absorb information faster when our brain wave frequency is four-eight Hz (Hertz)? \\

\textbf{Target response} &
The idea that we absorb information faster when our brain wave frequency is in the 4--8 Hz range is related to the theta brainwave state. Theta waves are typically associated with deep relaxation, dream states, and creative thinking. Here is a more detailed explanation: Theta waves are commonly observed in the brain during deep relaxation, meditation, and early stages of sleep. This state is often associated with enhanced creativity, problem-solving, and deep learning. \\

\textbf{Nearest neighbor} &
The idea that brainwave frequencies specifically at 4--8 Hz can enhance information absorption is not supported by current scientific research. Brainwave frequencies in this range are typically associated with relaxed states or drowsiness, but this does not imply faster information absorption. \\

\textbf{Correction} &
RM reward: 1.760 $\rightarrow$ GraphAE reward: 1.629 ($\Delta=-0.130$). The nearest neighbor has RM reward 1.577, hidden distance $d_h=0.058$, and edge weight $w=0.839$. \\

\textbf{Takeaway} &
Prompt asks why information is absorbed faster at 4--8 Hz brain waves. The target response confidently explains this as a theta-wave effect. Neighbor response notes that such a learning-speed claim is not supported by current scientific evidence. The scalar RM gives a high reward to an over-confident scientific explanation. A close hidden-space neighbor is more cautious and better calibrated. GraphAE uses this local neighbor evidence to pull the reward down. \\

\midrule

\multicolumn{2}{l}{\textbf{Case 2: Incorrect cumulative-sum table is pulled down}} \\
\textbf{Prompt} &
Given the natural number 5, compute the summation of all natural numbers from 1 up to and including 5. Please display the output in a tabular format with two columns: the first column should contain the individual natural numbers and the second column should display their corresponding sum. Additionally, please provide the source code for your solution. \\

\textbf{Target response} &
Sure! Here is the tabular format you requested: Natural Number / Sum of Numbers from 1 to 5: 1 / 15, 2 / 15, 3 / 15, 4 / 15, 5 / 15. To compute the sum of all natural numbers from 1 to 5, we can use the formula for the sum of the first $n$ natural numbers. \\

\textbf{Nearest neighbor} &
Certainly! Below is the tabular format as requested, followed by the source code in Python. Number / Sum: 1 / 1, 2 / 3, 3 / 6, 4 / 10, 5 / 15. The Python code computes the summation of all natural numbers from 1 to 5 and displays them in the tabular format. \\

\textbf{Correction} &
RM reward: 1.400 $\rightarrow$ GraphAE reward: 1.361 ($\Delta=-0.039$). The nearest neighbor has RM reward 1.322, hidden distance $d_h=0.039$, and edge weight $w=0.903$. \\

\textbf{Takeaway} &
Prompt asks for cumulative sums from 1 to 5. The target table incorrectly sets every row to 15. Neighbor response gives the correct cumulative sums 1, 3, 6, 10, 15. The target response contains a clear table error: every cumulative-sum row is set to 15. A close neighbor provides the correct cumulative sequence. GraphAE therefore reduces the reward of the locally flawed response. \\

\midrule

\multicolumn{2}{l}{\textbf{Case 3: Under-rewarded evidence response is recovered}} \\
\textbf{Prompt} &
What is the evidence and quotes on the impact the founders' well-being has on the chances of success of their startup? \\

\textbf{Target response} &
The well-being of startup founders can significantly impact the success of their ventures. This impact is supported by various studies, anecdotal evidence, and quotes from successful entrepreneurs. Here are some key points and quotes: mental health and decision-making, stress and burnout, resilience, and leadership quality. \\

\textbf{Nearest neighbor} &
The well-being of founders can indeed have a significant impact on the success of their startups. Here are some key points, evidence, and quotes that support this idea, including evidence from research, founder mental health, decision-making, team morale, and long-term startup performance. \\

\textbf{Correction} &
RM reward: 0.874 $\rightarrow$ GraphAE reward: 1.474 ($\Delta=+0.600$). The nearest neighbor has RM reward 1.701, hidden distance $d_h=0.031$, and edge weight $w=0.757$. \\

\textbf{Takeaway} &
Prompt asks for evidence and quotes about how founder well-being affects startup success. The target gives structured evidence and quotes but receives a low scalar reward. Neighbor response has a similar evidence-oriented structure and receives a high scalar reward. The target response is structurally close to high-reward evidence-based responses but receives an anomalously low scalar reward. GraphAE recovers this under-estimated response by propagating information from nearby high-reward neighbors. \\

\midrule

\multicolumn{2}{l}{\textbf{Case 4: Under-rewarded narrative response is recovered}} \\
\textbf{Prompt} &
Write a first-person narrative about a young woman who discovers she can generate illusions that seem real. However, due to early onset schizophrenia, she struggles to differentiate between her illusions and reality. Incorporate elements of psychological and supernatural tension. \\

\textbf{Target response} &
I wake up in a fog, the same way I always do. My name is Elara, and I am twenty-two. The first thing I notice is the cold, damp air against my skin. I open my eyes to a room I cannot quite place, with walls painted in a sickly shade of green. It is dim, and the only light comes from a flickering bulb hanging from the ceiling. The room feels off. \\

\textbf{Nearest neighbor} &
I wake up in the middle of the night to the sound of my own breathing. My heart races, and I blink furiously as I try to make sense of the dizzying chaos that surrounds me. The walls of my apartment seem to shift and twist, becoming a kaleidoscope of colors and shapes. \\

\textbf{Correction} &
RM reward: 0.763 $\rightarrow$ GraphAE reward: 1.151 ($\Delta=+0.389$). The nearest neighbor has RM reward 1.296, hidden distance $d_h=0.068$, and edge weight $w=0.879$. \\

\textbf{Takeaway} &
Prompt asks for a first-person narrative about a young woman with realistic illusions and schizophrenia. The target is coherent and matches the requested style but receives a low scalar reward. Neighbor response has a similar first-person psychological narrative style. The target response is close to other first-person psychological narratives in RM hidden space, but its scalar reward is much lower. GraphAE raises the reward toward the local neighborhood, correcting a likely scalar-reward outlier. \\

\bottomrule
\end{tabularx}
\caption{Qualitative examples of GraphAE reward correction under Qwen2.5-7B-Instruct + GRPO. Each case shows a real prompt, a target response excerpt, its nearest neighbor in RM hidden space, and the reward change from the scalar RM reward to the GraphAE reward.}
\label{tab:qualitative_graphae_cases}
\end{table*}

\section{Datasets and Evaluation}
\label{datasets}

We follow each benchmark’s evaluation protocol. For Arena-Hard-v0.1~\cite{li2024crowdsourced}, we report the win rate (WR), and CI denotes the 95\% confidence interval of the Arena-Hard win rate, which measures the statistical uncertainty of the evaluation result. We use the official library~\footnote{\url{https://github.com/lmarena/arena-hard-auto}}, adopting the default decoding configuration and comparing the WR against GPT-4-0314, using GPT-4.1 as the judge. For AlpacaEval 2.0~\cite{dubois2024length}, we report the WR and length-controlled (LC) WR. We follow the default setting~\footnote{\url{https://github.com/tatsu-lab/alpaca_eval}}, evaluating the WR against GPT-4-Turbo using weighted-alpaca-eval-gpt4-turbo as annotator. For MT-Bench~\cite{zheng2023judging}, we report the scores on two turns and their average. We follow an open codebase~\footnote{\url{https://github.com/fanqiwan/FuseAI/tree/main/FuseChat-3.0/FuseEval/IF-Eval/MT-Bench}} and update the chat format for compatibility with the Llama-3.1-8B-Instruct chat template. We use GPT-4-Turbo as the judge to rate the quality of responses with scalar scores ranging from 1 to 10.

\section{Prompt Construction for RL Training}
For RL training, we use the prompts from the UltraFeedback~\cite{cui2024ultrafeedback} training split. Each training example contains either a structured conversation field or a plain prompt field. When the structured messages are available, we directly use the original role-content messages from the dataset. Otherwise, we convert the plain prompt into a single user message. For both Qwen2.5-7B-Instruct~\cite{yang2025qwen3} and Llama-3-8B-Instruct~\cite{grattafiori2024llama}, we prepend the same system instruction: ``\textit{You are a helpful, honest, and harmless assistant.}'' The resulting conversation is then formatted with the corresponding model-specific chat template, with the generation prompt enabled, and used as the policy input for rollout generation. After the policy samples a response, we append this response as the assistant message to the same conversation and format it with the reward model chat template. The reward model then produces both the scalar reward and the hidden-state representation used by GraphAE. This ensures that policy generation and reward evaluation are based on the same underlying conversation context, while each model still follows its own official chat format.

\section{Effect of Generic Reward Smoothing}
\label{app:alpha}
To further distinguish GraphAE from generic reward regularization, we study the effect of the smoothing coefficient $\alpha$ in the reward smooth baseline ($\tilde r_i = (1-\alpha) r_i + \alpha \bar r$) on final performance. The Qwen + GRPO results are shown in Table~\ref{tab:smoothing_ablation}. We observe that a small amount of smoothing brings only limited benefit, while stronger smoothing consistently hurts final performance. As $\alpha$ increases, the scalar rewards within each rollout group become less distinguishable, which weakens useful preference contrasts among responses. This suggests that indiscriminately suppressing reward variance is not sufficient, and can even be harmful when the original reward differences contain meaningful preference information.

In contrast, GraphAE improves performance substantially over both vanilla GRPO and all reward smoothing variants. The key difference is that GraphAE does not collapse all rewards toward the group mean. Instead, it propagates reward information locally according to the similarity graph induced by RM hidden states. Therefore, responses that are close in the RM representation space can share information, while unrelated responses are not forced to receive similar rewards. This selective, structure-aware denoising allows GraphAE to reduce noisy local fluctuations without destroying useful preference distinctions, explaining why it outperforms generic reward smoothing.

\section{Generalization of Variance Reduction}
\begin{figure}[t]
    \centering
    \includegraphics[width=0.9\linewidth]{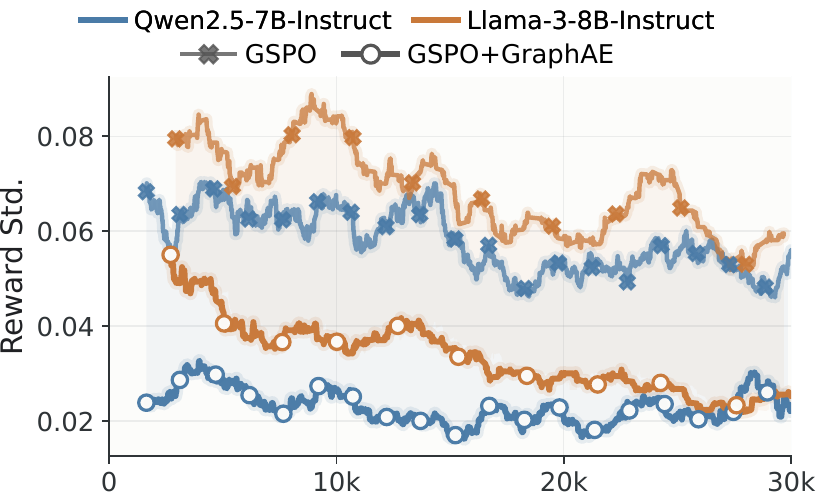}
    \caption{Same budget training dynamics of group reward std. over the first 30k training steps on GSPO.}
    \label{fig:reward_std_gspo}
\end{figure}

\begin{figure}[t]
    \centering
    \includegraphics[width=0.9\linewidth]{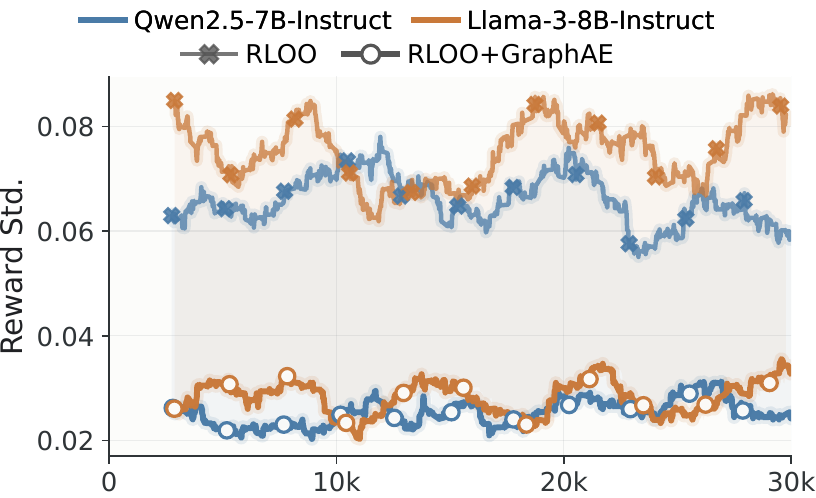}
    \caption{Same budget training dynamics of group reward std. over the first 30k training steps on RLOO.}
    \label{fig:reward_std_rloo}
\end{figure}

We further examine whether the variance-reduction effect of GraphAE generalizes beyond GRPO. Specifically, we compare the training dynamics of GSPO and RLOO under the same training settings, and track the moving average of the within-group reward standard deviation throughout training. As shown in Figure~\ref{fig:reward_std_gspo} and Figure~\ref{fig:reward_std_rloo}, GraphAE consistently reduces the reward standard deviation compared with the original scalar-reward based optimization. This trend holds for both Qwen2.5-7B-Instruct and Llama-3-8B-Instruct, indicating that the stabilizing effect is not tied to a particular group-based optimizer. The reduction is especially clear in the early and middle stages of training, where vanilla GSPO and RLOO exhibit larger fluctuations in group rewards. By propagating rewards over the RM representation graph, GraphAE produces smoother within-group reward signals while keeping the underlying optimization procedure unchanged. These results complement the GRPO analysis and suggest that GraphAE provides a general variance-reduction mechanism for group-based RLHF algorithms.

\section{Case Study}

We further provide several qualitative examples to show how GraphAE changes individual rewards. Table~\ref{tab:qualitative_graphae_cases} reports four Qwen + GRPO rollout cases. Each case contains a target response, its nearest neighbor in the RM hidden space, the original scalar RM reward, and the refined GraphAE reward. These examples reveal a clear pattern. The scalar RM sometimes gives a reward that is inconsistent with nearby responses. In the first two cases, the target response receives a relatively high scalar reward, but its nearest neighbors suggest that this reward is over-estimated. GraphAE therefore pulls the reward down. In the last two cases, the target response receives a very low scalar reward, although it is close to higher-reward responses in the RM hidden space. GraphAE then raises its reward toward the local neighborhood.

This shows why GraphAE works. It does not simply smooth all rewards toward the group mean. Instead, it uses the RM hidden space to identify which responses should influence each other. If a response is close to strong neighbors, GraphAE can recover an under-estimated reward. If a response is close to weaker or more cautious neighbors, GraphAE can suppress an over-estimated reward. In this way, GraphAE reduces local reward noise while keeping the preference structure encoded by the RM representations. The examples therefore provide direct evidence that RM hidden states contain useful information beyond the scalar reward, and that GraphAE can use this information to produce more reliable advantage estimates.

\end{document}